\documentclass{article}

\PassOptionsToPackage{numbers, compress}{natbib}

\usepackage[final]{neurips_2024}

\usepackage[utf8]{inputenc} 
\usepackage[T1]{fontenc}    
\usepackage{hyperref}       
\usepackage{url}            
\usepackage{booktabs}       
\usepackage{amsfonts}       
\usepackage{nicefrac}       
\usepackage{microtype}      
\usepackage{xcolor}         

\usepackage{lipsum}
\usepackage[toc,page]{appendix}
\usepackage{amsmath}
\usepackage{cleveref}
\usepackage{multirow}
\usepackage{comment}
\usepackage{graphicx}

\title{Neural machine translation of clinical procedure codes for medical diagnosis and uncertainty quantification}

\author{%
  Pei-Hung Chung \\
  Department of Electrical and Computer Engineering\\
  Texas A\&M University\\
  College Station, TX 77843 \\
  \texttt{chung95191@tamu.edu} \\
  \And
  Shuhan He$^*$\\
  Emergency Medicine \\
  Massachusetts General Hospital \\
  Boston, MA 02115 \\
  \texttt{she@mgh.harvard.edu} \\
  \And
  Norawit Kijpaisalratana \\
  Emergency Medicine \\
  Massachusetts General Hospital \\
  Boston, MA 02115 \\
  \texttt{nkijpaisalratana1@mgh.harvard.edu} \\
  \And
  Abdel-badih el Ariss \\
  Emergency Medicine \\
  Massachusetts General Hospital \\
  Boston, MA 02115 \\
  \texttt{aariss@mgh.harvard.edu} \\
  \And
  Byung-Jun Yoon\thanks{Corresponding authors}\\
  Computational Science Initiative \\
  Brookhaven National Laboratory \\
  Upton, NY 11973 \\ [3pt]
  Department of Electrical and Computer Engineering\\
  Texas A\&M University\\
  College Station, TX 77843 \\
  \texttt{bjyoon@tamu.edu} \\
}

\begin{document}

\maketitle
\begin{abstract}
  A Clinical Decision Support System (CDSS) is designed to enhance clinician decision-making by combining system-generated recommendations with medical expertise. Given the high costs, intensive labor, and time-sensitive nature of medical treatments, there is a pressing need for efficient decision support, especially in complex emergency scenarios. In these scenarios, where information can be limited, an advanced CDSS framework that leverages AI (artificial intelligence) models to effectively reduce diagnostic uncertainty has utility. Such an AI-enabled CDSS framework with quantified uncertainty promises to be practical and beneficial in the demanding context of real-world medical care. In this study, we introduce the concept of Medical Entropy, quantifying uncertainties in patient outcomes predicted by neural machine translation based on the ICD-9 code of procedures. Our experimental results not only show strong correlations between procedure and diagnosis sequences based on the simple ICD-9 code but also demonstrate the promising capacity to model trends of uncertainties during hospitalizations through a data-driven approach.
\end{abstract}

\section{Introduction}

When a patient presents for clinical care hospital, clinicians sometimes face the challenge of initial uncertainty\cite{alam2017managing}, necessitating more data from examinations and observations. This uncertainty is often highest at the point of first presentation and can be addressed through a series of appropriate procedures. As treatment progresses, effective therapies can progressively reduce the unknown aspects of the patient's condition. The goal is to minimize this uncertainty with effective treatments, as each illness presents several viable treatment options. However, the evolving nature of the patient's condition demands the identification of the most suitable treatment plan. In this context, a Clinical Decision Support System (CDSS) becomes crucial, assisting clinicians in decision-making by efficiently narrowing down uncertainties with limited information\cite{sutton2020overview}.

In this evolving landscape of clinical decision-making, the use of CDSS becomes pivotal. CDSSs help clinicians navigate through the complexities of medical care, much like Automatic Speech Recognition (ASR)\cite{benzeghiba2007automatic} systems use phonemes as the minimal units to represent their semantic contents from the speech signals in the acoustic models, then predict the combinations of plausible words in sentences in the language models. In this study, we introduce a novel framework centered around reducing uncertainty in clinical decision-making. The initial validation of this framework is conducted through a retrospective review utilizing the International Classification of Diseases-Ninth Revision (ICD-9) and Current Procedural Terminology (CPT) procedure codes. We emphasize that the use of ICD-9 and CPT codes is instrumental for validation purposes, serving to corroborate our primary approach towards entropy and uncertainty quantification, and ultimately, the reduction thereof. While these codes are typically generated post patient stay, thus not providing prospective data, their utility in affirming the validity of our uncertainty quantification/reduction framework is invaluable. Our approach in this study harnesses the systematic structure of ICD-9 and CPT codes to encode the potential patients’ conditions and further predict subsequent steps in a sequence, employing them like the elements of phoneme and subword in ASR system to guide decision-making. This methodology enables CDSSs to efficiently narrow down uncertainties with limited initial information, thereby assisting clinicians in making informed decisions\cite{sutton2020overview}. Furthermore, the design of a CDSS integrates these encoded guidelines with the clinician’s medical expertise, forming a synergy that is crucial in time-critical and resource-intensive medical scenarios. While standard operating procedures exist for inpatients, the dynamic and often critical nature of hospital environments demands quick yet accurate decision-making. Here, the CDSS, empowered by its encoded data akin to a linguistic system, plays a vital role in guiding clinicians through complex medical situations.

CDSS can be classified as knowledge-based and non-knowledge based. For knowledge-based systems, decisions are made based on predefined rules and medical guidelines\cite{kabachinski2013look}. In contrast, non-knowledge-based systems facilitate physicians making precise arrangements by data-driven approaches based on artificial intelligence (AI) / machine learning (ML) models \cite{antoniadi2021current}. Knowledge-based systems achieve great success in diagnostic support when giving medical expertise on symptoms and side effects. As for AI/ML models, despite their remarkable capacity to manipulate massive amounts of data in non-knowledge-based systems, they are a black box\cite{xu2023interpretability}. Data availability might also restrict feature extraction from provided data to represent the patient's health condition\cite{chen2022barriers}. In other words, features obtained from vital signs, electrocardiograms, or laboratory results during a hospital stay might not be viable when a patient is newly admitted to the hospital or when medical facilities are not applicable for any reason.     

Information entropy, also known as Shannon entropy\cite{shannon1948mathematical}, has demonstrated its utility in digital communication and data compression\cite{ash2012information}. This idea of information entropy enables us to quantify the amount of uncertainty (or variability) of quantities of interest (QoI) based on their probability distributions and it has been also playing a central role in AI/ML.

Medical entropy has been previously applied to clinical calculators and has been proposed as a substitute for sensitivity and specificity\cite{he2024entropy}. In this study, we propose a framework to measure medical uncertainty at every stage during specific admissions by estimating the uncertainties caused by heterogeneous factors such as medical history, multifarious etiology, or uncaptured data in medical scenarios. This aims to optimize clinical decision-making by providing a more nuanced and comprehensive understanding of patient-specific variables, ultimately leading to tailored, efficient, and effective patient care through AI-enabled CDSS.

\section{Materials and Methods}

\subsection{Data Source and Experiment Compute Resource}

In this study, we focused on the Medical Information Mart for Intensive Care (MIMIC)-IV database\cite{johnson2023mimic}, particularly on the International Classification of Diseases (ICD) codes used for diagnosing and documenting procedures during hospital admissions. Our experiments were conducted using cases recorded with ICD-9 codes\cite{world1988international}, encompassing a total of 155,933 admissions. 

ICD codes provide a standardized framework for diagnosis, while CPT codes document the procedures carried out. Although these codes may not directly mirror actual medical practices, they offer a clear and structured way to understand the activities and decisions during a patient's hospital stay\cite{henderson2006quality}. This structured approach is not only beneficial for clarity in clinical understanding but also proves invaluable for AI/ML  applications. The use of ICD codes for encoding diagnostic terms in clinical documentation has been a focal point of research, underscoring their crucial role in efficiently extracting and analyzing essential data from electronic health records\cite{atutxa2017machine}. This synergy between standardized medical coding and AI/ML is a key aspect of our study.

All experiments in this study were conducted on a CPU, using an Apple MacBook Pro equipped with the M1 Max chip. This setup not only provided sufficient computational power for processing the ICD-9 coded data and running the machine learning models, but also enabled us to carry out the experiments with relative ease, without requiring additional high-performance computing resources such as GPUs or cloud-based servers. This also ensured that the study remains reproducible on accessible hardware for users with similar computational setups.

\subsection{Model Frameworks}

\begin{figure*}[t]
  \centering
  \includegraphics[width=\linewidth]{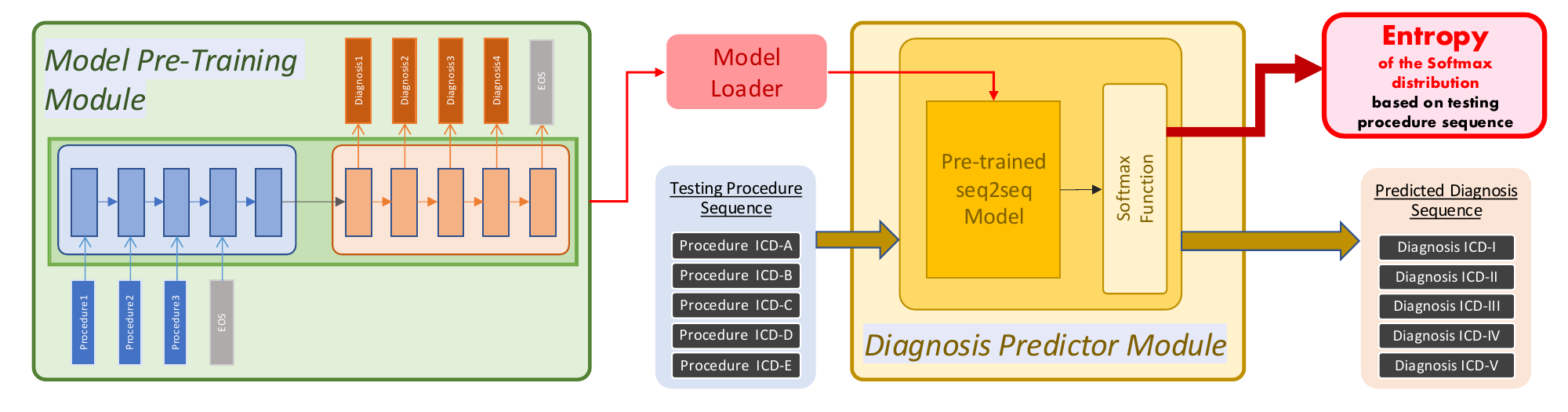}
  \caption{The proposed framework of entropy quantification includes Model Pre-Training Module (MPM) and Diagnosis Predictor Module (DPM) by using the procedure and diagnosis ICD-9 codes in the MIMIC-IV database.}
  \label{fig:framework}
\end{figure*}

Figure~\ref{fig:framework} illustrates the proposed framework of entropy quantification during hospitalization, which consists of two main modules: the Model Pre-training Module (MPM) and the Diagnosis Predictor Module (DPM). To specify, the upstream MPM is for pre-training predictive models, and the downstream DPM takes the pre-trained model to obtain the output distribution while predicting diagnosis. Given the testing procedure sequence, which can be arbitrary, the decoder in DPM can further predict diagnosis. Meanwhile, we can obtain the entropy of the output distribution, which is also the confidence of potential diagnosis.

In our framework's application to a clinical setting, the process begins when a physician inputs patient data accumulated since admission into the DPM. This model evaluates the data to estimate the diagnostic likelihood, represented as a probability distribution across potential diagnoses. It's important to note that these diagnoses represent a combination of different health conditions at a specific stage, rather than a single disease state. 

Subsequently, the DPM calculates the information entropy derived from this diagnostic probability distribution, presenting it as “medical entropy” to the clinician. This entropy serves as a quantitative measure of uncertainty in the diagnosis. The physician, upon reviewing the medical entropy, can then propose potential interventions for the subsequent stage of patient care, aiming at reducing this uncertainty. This action allows the clinician to assess how these interventions might increase or decrease the medical entropy, thereby refining or expanding the differential diagnosis.
 
Throughout this interactive process, the physician gains insights not only into the immediate medical entropy linked to the current clinical query but also into a sequence of potential therapeutic options, each with their respective posterior medical entropies. Moreover, by inputting various stages of patient care into the system, the physician can analyze the trajectory of medical entropy, identifying key interventions that substantially impact patient outcomes. This aspect of the framework is particularly beneficial for understanding the dynamic nature of medical decision-making over the course of a patient's hospital stay.

To make this framework workable in general use, we utilize the procedure sequences and diagnosis sequences in the form of ICD codes as the observations in the hospital and the diagnoses of the current stage. In this model, the sequence of procedure-related ICD codes constitutes the input for the MPM, and the sequence of diagnostic ICD codes forms its output. Notably, this does not mean that the proposed framework cannot apply features such as vital signs or lab results to the MPM. Instead, the main reason we adopt the plain ICD codes as training data instead of utilizing the more complex feature embedded in more information is that we focus on keeping the framework flexible and workable in medical practice. 

To address the issue of the capacity of the predictive model in MPM, we apply non-knowledge-based decision-making for the CDSS strategy. Still, this predictive model can be substituted with a different model architecture, should it be better suited for other settings. Since the source and target data are in sequence, in the current study, we exploit the seq2seq models\cite{sutskever2014sequence} as the diagnosis predictor in MPM to duplicate a success for the tasks in the fields of natural language processing (NLP)\citep{chowdhary2020natural,khurana2023natural,stahlberg2020neural,yousuf2021systematic}.

\subsection{Model Overview}

\begin{figure*}[t]
  \centering
  \includegraphics[width=\linewidth]{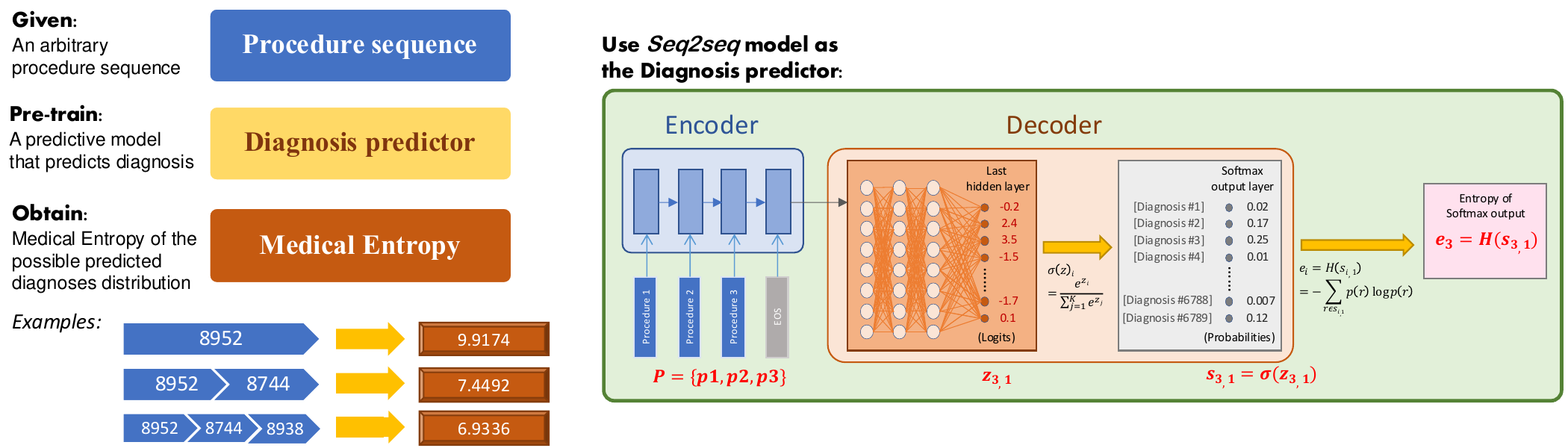}
  \caption{Illustration of an implementation of the proposed model by adopting the seq2seq model as the pre-trained model and an example of the entropy quantification of a procedure sequence. }
  \label{fig:EQ}
\end{figure*}

Figure~\ref{fig:EQ} illustrates the models we apply to implement the proposed framework and the examples of obtaining the entropy trend from a procedure sequence. In this study, we utilize the seq2seq model to pre-train the model for diagnosis predictor. To construct the entropy trend of a procedure sequence we desired, we feed the procedures to the encoder in order, then the decoder will output a distribution of possible diagnoses. The distribution here is determined by the softmax function of the last hidden layer of the seq2seq diagnosis predictor. Eventually, we then obtain the entropy based on a specific procedure combination. Take the procedure sequence 8952-8744-8938 as an example, we first obtain an entropy of 9.9174 by taking {8952} as the input of the diagnosis predictor. Further, we get the entropies of 7.4492 and 6.9336 by feeding the procedure combinations {8952, 8744} and {8952, 8744, 8938}, respectively.

We adopt the seq2seq architecture model to predict suggestive diagnoses based on the procedures received so far due to its similar characteristics to data in Natural Language Processing (NLP)-related tasks. The data in both fields are in sequence and have varied lengths. However, there are still major differences between the two. Each word in a source sentence could be crucial for predicting the target sequence in machine translation. In contrast, the importance of a specific diagnosis code in diagnosis sequences ranks by its order. Moreover, the procedure code can possess multiple significances for different orders, combinations, and repetition frequencies. 

On the other hand, the source and the target sequences in NLP-related tasks have vital cause-and-effect relationships. On the contrary, it does not work for the procedure-diagnosis relationship. Namely, the diagnosis for an admission can derive from the given procedures that a patient had received at a specific moment. Yet, the bond between procedure and diagnosis becomes insignificant when reasoning backward since identical procedure sequences can lead to various diagnosis combinations. This adverse impact could be amplified if admissions with a single procedure are the vast majority in the dataset. Due to the insufficiently informative input as a single procedure, it is explicit that it would be unreasonable for the model to predict the diagnosis with ample information.

In the admissions with the ICD-9 code we adopt in the MIMIC-IV dataset, there are 57,322 and 36,040 cases for the admissions with one and two procedures, respectively. This includes 59.9\% of the admissions in total. Take ICD code "9925" as an example, there are 1,947 admissions among admissions with a single procedure. The code "9925" entails "Injection or infusion of cancer chemotherapeutic substance," which means that we know that the patients come to the hospital for chemotherapy. It is reasonable that the patients receive only one procedure; however, the diagnoses for these cases are widely different. 

\subsection{Seq2seq diagnoses predictor}

To predict diagnoses based on procedures during patient admission, we use a Seq2Seq model. Both diagnosis and procedure codes are sequential, with procedure codes following a chronological order, while the order of diagnosis codes reflects the severity of the patient's condition. The Seq2Seq model offers a flexible approach to capture this dynamic by processing variable-length sequences, making it ideal for linking procedures to diagnoses.

The Seq2Seq framework, based on the encoder-decoder architecture \cite{sutskever2014sequence}, takes a source sequence as input and generates a target sequence as output. The encoder processes the input sequence to create a context vector, an embedding of the entire source sequence, which is then passed to the decoder to generate the target sequence. In our case, the encoder processes the procedure sequence, and its final hidden state serves as the context vector. The decoder uses this context to predict diagnoses sequentially, with each prediction informing the next.

We also integrate an attention mechanism \citep{luong2015effective, bahdanau2014neural} to allow the decoder to weigh and focus on the most relevant hidden states from the encoder, enhancing prediction accuracy. To further improve training efficiency and prevent error propagation, we apply the teacher-forcing technique \cite{belousov2019entropic}, which inputs the ground truth directly during training.

We focus on predicting the distribution of diagnoses at different stages of admission using ICD-9 codes. The encoder is fed with procedure ICD-9 codes, while the decoder outputs diagnosis ICD-9 codes. During evaluation, the softmax output of the decoder provides a probabilistic distribution of diagnoses, aiding in entropy-based analyses.

\subsection{Entropy Quantification in Admissions}

For the downstream task of the diagnoses predictor, we now focus on the input procedures and decoder's softmax output of the pre-trained model. Given a procedure sequence $P = \{p_1,p_2,...,p_M\}$  with $M$ procedures during an admission, we have the predicted diagnosis sequence $D_m = \{d_{m,1}, d_{m,2},..., d_{m,n}\}$ for every single step $m$ in procedure sequence $P$, where $m = 1, …, M$. For every diagnosis code $d$ in the predicted diagnosis sequence $D$, the decoder outputs the diagnosis based on the softmax function $s_{m,k}$ of the last neural layer $z_{m,n}$ in the deep neural network of the decoder, where

\begin{equation}
    s_{m,n} = \sigma (z_{m,n}), \quad and 
\end{equation}

\begin{equation}
    \sigma (z)_i = \frac{e^{z_i}}{\sum_{j=1}^{K} e^{z_j}} \quad for  \  i=1,...,K \quad \textrm{and} \quad z = (z_1,...,z_K) \in \mathbb{R}^K
\end{equation}

$s_{m,k}$ is a vector that signifies the confidence of all the potential diagnoses candidates. The dimension of the $s_{m,n}$ is 6789 as there are 6789 codes in the dictionary of diagnoses ICD-9 codes in the MIMIC-IV database. Here, we choose the corresponding softmax output $s_{m,1}$ of the very first predicted diagnosis $d_{m,1}$, representing the probability distribution of all potential outcomes (i.e., diagnosis code). This means that the softmax output $s_{m,1}$ is directly determined by the context vector c that encodes all procedures in order, which is used by the decoder to predict the most important diagnosis code.

As a consequence, we obtain the sequence of softmax output $S = \{s_{1,1}, s_{2,1}, ..., s_{M,1}\}$ as the confidence of possible diagnoses by given cumulative input procedure sequence $CP_m = \{p_1, p_2,...,p_m\}$. During the hospitalization of a given patient, from admission to discharge, we update the corresponding distribution after receiving every procedure. To further analyze the trends of the distributions, we quantify the uncertainty of the predicted diagnosis code by calculating the information entropy of this distribution. In this way, we have the entropy sequence $E = \{e_1, e_2, ..., e_m\}$, where

\begin{equation}
e_i = H(s_{i,1}) = -\sum_{r \in s_{i,1}} p(r) \ \textrm{log}\ p(r) = \mathbb{E} [- \textrm{log} \ p(s_{i,1})]
\end{equation}

Despite the semantic differences between the diagnosis-procedure data and NLP-related data, we assume that the seq2seq model still works as a diagnosis predictor since the diagnosis-procedure settings share similar sequential behavior with typical seq2seq scenarios in NLP tasks. To ameliorate the error propagation between the pre-trained model and the downstream entropy quantification, it is necessary to expose how satisfyingly the diagnosis predictor could go among different model architectures. Hence, defining the pertinent and reliable evaluation metrics for the diagnosis predictor is crucial and needed.

\subsection{Model Performance Evaluation}

Various evaluation metrics are defined in the field of NLP to assess diverse aspects. However, widely used evaluation metrics for the NLP area may not necessarily be suitable in our scenario. This is because, unlike typical NLP models, the diagnosis predictor cares less about the N-gram terms of the predicted diagnosis sequence. Many NLP evaluation metrics have a penalty term to ensure the model decodes longer sequences, which is not applicable in our case. Moreover, the order of the diagnosis sequence has imperative information about how important it is to the patient's condition, which is not the case in a natural language sequence.

For reasons stated above, instead of utilizing metrics such as the Word error rate (WER), Bilingual Evaluation Understudy (BLEU)\cite{papineni2002bleu}, or their variants\cite{lin2004rouge}, we utilize the f1 score and Jaccard index\cite{costa2021further} to examine the model performance of the seq2seq diagnosis predictor. Furthermore, we also propose First-N-accuracy to assess the model's capability of precisely predicting the most decisive diagnoses. That is to say, if N equals three, the First-N-accuracy will be counting the percentage of the first three predicted diagnoses appearing in the first three diagnoses in the ground truth.    

To further assess the model's ability to handle uncertainty, we focus on the entropy trend throughout entire admissions from a global perspective. The dataset's procedure and diagnosis sequence pairs, reflecting real-world medical decisions, serve as ideal cases for examining entropy reduction. Every action by physicians is considered an attempt at reducing entropy. Despite potential fluctuations in individual entropy trends due to the unpredictable nature of hospital scenarios, the overall entropy trend should consistently decrease, regardless of the patient outcomes, be it discharge or passing away. This approach allows us to evaluate the model's performance in a dynamic, real-world medical setting.

\section{Results}

\begin{table}[]
\centering
\caption{Model performance of Diagnosis Predictor as F1-score, Jaccard Index, and First-N-Accuracy among different seq2seq model architectures.}
\label{tab:model_performance}
\scalebox{0.7}{
\begin{tabular}{@{}cclllll@{}}
\toprule
\multicolumn{2}{c}{\multirow{2}{*}{\begin{tabular}[c]{@{}c@{}}Attention-based \\ seq2seq Diagnosis Predictor\end{tabular}}} &
  \multicolumn{1}{c}{\multirow{2}{*}{F1-score}} &
  \multicolumn{1}{c}{\multirow{2}{*}{\begin{tabular}[c]{@{}c@{}}Jaccard \\ Index\end{tabular}}} &
  \multicolumn{3}{c}{First-N-Accuracy} \\ \cmidrule(l){5-7} 
\multicolumn{2}{c}{} &
  \multicolumn{1}{c}{} &
  \multicolumn{1}{c}{} &
  \multicolumn{1}{c}{N=1} &
  \multicolumn{1}{c}{N=2} &
  \multicolumn{1}{c}{N=3} \\ \midrule
\multirow{3}{*}{\begin{tabular}[c]{@{}c@{}}no\\ teacher forcing\end{tabular}} &
  1-layer &
  \textbf{1.08e-02} &
  \textbf{5.45e-03} &
  0.2717 &
  0.2236 &
  0.1951 \\
 &
  2-layer &
  9.94e-03 &
  5.00e-03 &
  0.2488 &
  0.2019 &
  0.1766 \\
 &
  3-layer &
  6.35e-03 &
  3.18e-03 &
  0.1871 &
  0.1669 &
  0.1512 \\ \midrule
\multirow{3}{*}{\begin{tabular}[c]{@{}c@{}}with\\ teacher forcing\end{tabular}} &
  1-layer &
  8.66e-03 &
  4.35e-03 &
  \textbf{0.2884} &
  \textbf{0.2281} &
  \textbf{0.2025} \\
 &
  2-layer &
  8.02e-03 &
  4.02e-03 &
  0.2321 &
  0.1986 &
  0.1847 \\
 &
  3-layer &
  3.14e-03 &
  1.57e-03 &
  0.1505 &
  0.1440 &
  0.1341 \\ \bottomrule
\end{tabular}
}
\end{table}

Table~\ref{tab:model_performance} summarizes the performance of the proposed framework for various model architectures. As can be seen in Table~\ref{tab:model_performance}, the simplest 1-layer model surpasses other models in performance across all the metrics used. A noteworthy observation is that the simpler 2-layer model significantly outperforms the 3-layer model, while the performance gap between the 1-layer model and the 2-layer model is relatively modest. Additionally, to compare the performance of these three model architectures with and without teacher-forcing to investigate its impact on the overall performance. We trained the models with identical hyperparameters, where the only difference was whether the teacher-forcing feature was used or not\cite{lamb2016professor}. The outcomes revealed that the models with teacher-forcing consistently mirrored the performance trends of models without teacher forcing. Furthermore, it was observed that the addition of teacher forcing did not enhance the performance compared to the original models in the tested scenarios.

\begin{figure*}[t]
  \centering
  \includegraphics[width=\linewidth]{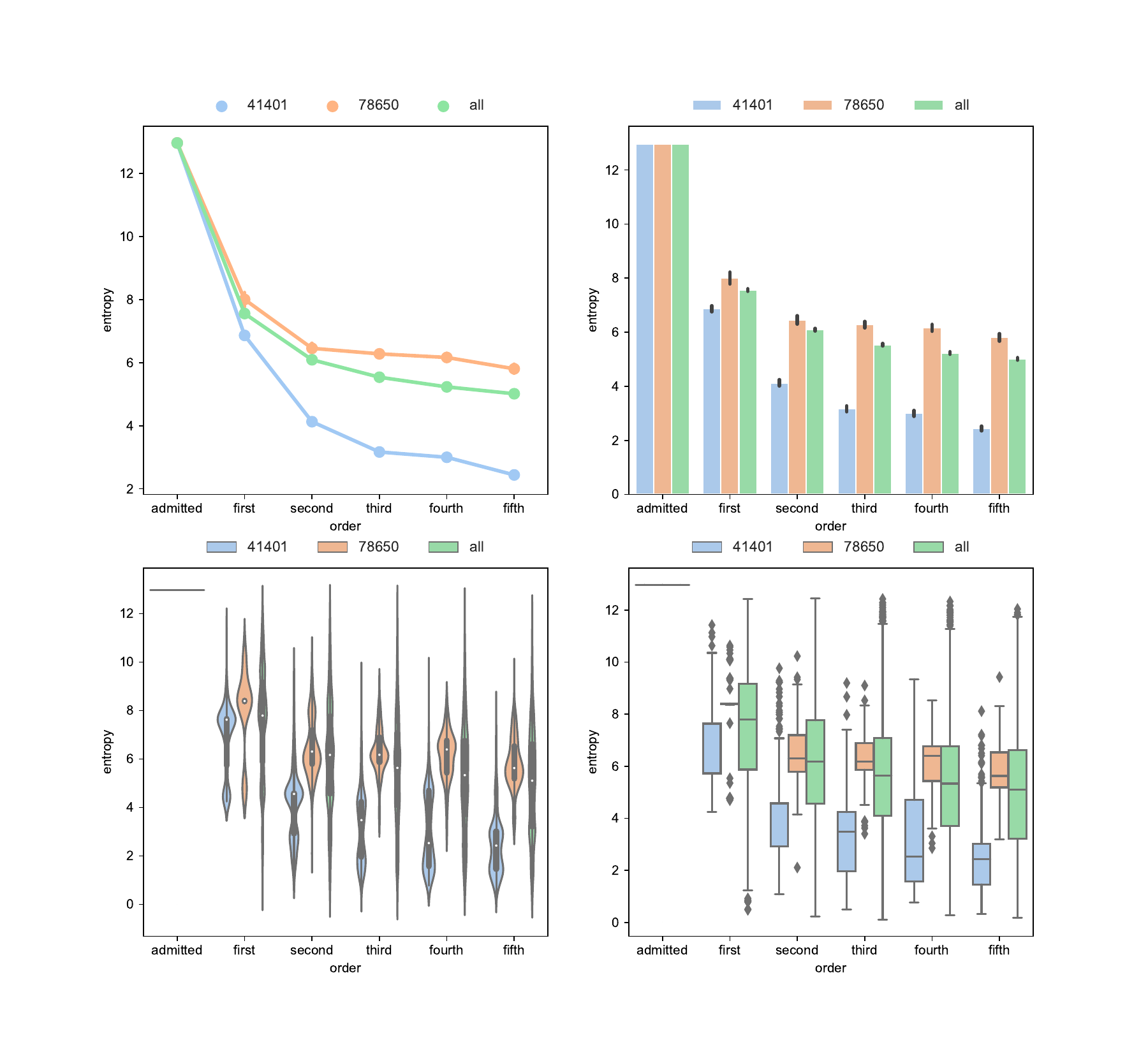}
  \caption{Trends of average entropy for admission cases with five procedures. The three colors show the entropy trends clustered by their primal diagnosis, which are diagnosis ICD-9 code 41401 (Coronary atherosclerosis of native coronary artery), 78650 (Chest pain), and all admissions with 5 procedures.}
  \label{fig:point_l5}
\end{figure*}

Figure~\ref{fig:point_l5} shows the average entropy trend of 3 different clusters, grouped by their primal diagnosis ICD codes. To better grasp the trends of uncertainty drops, we focus on the cases with the same length. Specifically, we selected the cases of patients who received a total of five procedures during their admissions. In the figure, "All" indicates the average entropy trends for all diseases, whereas "41401" and "78650" are groups clustered by their primary diagnosis, "Coronary atherosclerosis of native coronary artery" and "Chest pain," respectively. The total cases among the three clusters, All, 41401, and 78650, are 8,705, 679, and 220 admissions, respectively.

As shown in Figure~\ref{fig:point_l5}, all the admissions have the same initial entropies at the very beginning of the admissions. That is the condition when the patients arrive at the hospital without receiving any procedure. To provide a reference for comparisons, we assume that the  6,984 potential procedures are equally distributed. In this case, the initial entropy is set to 12.76. In another case, this initial entropy will be 9.43 if we assume the potential procedures are distributed by their frequencies (estimated based on the MIMIC-IV dataset). In all the figures in this paper, we set the initial entropy to 12.76, taking that the initial distribution of procedures is uniform.

Note that the "entropy trend" mentioned in this paper reveals the entropy of the distributions of the potential diagnoses when receiving a new procedure made by the physician during the hospitalization. The entropy trend reflects how the medical entropy of patients changes from admission to discharge, as a result of receiving procedures. If the patient received three procedures during the entire hospitalization, for example, procedures $A$, $B$, and $C$, we would have four entropies in their entropy trend. The entropy trend includes the initial entropy without receiving any procedure and the respective entropies of the predicted diagnosis distributions by given procedure sequence $\{A\}$, $\{A, B\}$, and $\{A, B, C\}$. Overall, Figure~\ref{fig:point_l5} clearly shows that the average entropy tends to decrease as the number of received procedures increases for all three clusters.

Next, we investigated the entropy drop for the most frequent first N procedures (for N=1, 2, and 3) since the very start of admission. These results are summarized in~\cref{tab:1-gram,tab:2-gram,tab:3-gram}. Information theory states that the more information we know about the world (or the patient’s state, in this study), the smaller entropy we get as a result (i.e., less uncertainty regarding the patient’s diagnosis in the current study). The drops of the entropies reveal the reduction of the uncertainties. As Figure~\ref{fig:point_l5} shows, the entropy tends to decrease when every upcoming procedure arrives, it is imperative to examine how entropy reduction acts among various procedures.  

Table~\ref{tab:1-gram} demonstrates the entropy drops of the ten most frequent first procedures at the beginning of the admission. The ICD-9 procedure code "7569"(ranked \#5) has the lowest entropy of 6.37, describing "Repair of other current obstetric laceration." In contrast, the ICD-9 procedure code "9671"(ranked \#9) has the highest entropy of 10.27 for "Continuous invasive mechanical ventilation for less than 96 consecutive hours."

\begin{table}[]
\centering
\caption{Entropy of most frequent first N=1 procedure.}
\label{tab:1-gram}
\scalebox{0.6}{
\begin{tabular}{@{}lcccc@{}}
\toprule
\multirow{2}{*}{\#} & \multirow{2}{*}{\textbf{ICD-9 code}} & \multirow{2}{*}{\textbf{Cases}} & \multirow{2}{*}{\textbf{Frequency}} & \textbf{Entropy After Receiving} \\
   &      &      &        & 1st procedure \\ \midrule
1  & 66   & 4528 & 2.90\% & 9.8625        \\
2  & 8938 & 4036 & 2.59\% & 7.0969        \\
3  & 741  & 3443 & 2.21\% & 7.6941        \\
4  & 8952 & 3426 & 2.20\% & 9.9164        \\
5  & 7569 & 3399 & 2.18\% & 6.3773        \\
6  & 3893 & 3242 & 2.08\% & 9.1968        \\
7  & 9925 & 3111 & 1.99\% & 7.2922        \\
8  & 3995 & 3019 & 1.94\% & 8.4310        \\
9  & 9671 & 2790 & 1.79\% & 10.2779       \\
10 & 5491 & 2703 & 1.73\% & 8.4703        \\ \bottomrule
\end{tabular}
}
\end{table}

To make the proposed framework applicable to real-world medical use, the entropy quantification of a specific procedure sequence in any situation has to be more explainable and informative to physicians instead of being a "black box". In other words, it is indispensable to analyze the entropy trends individually. For better demonstration, we choose examples of actual cases in the MIMIC-IV dataset with the same number of total received procedures and with similar final diagnoses.

\begin{figure*}[t]
  \centering
  \includegraphics[width=0.8\linewidth]{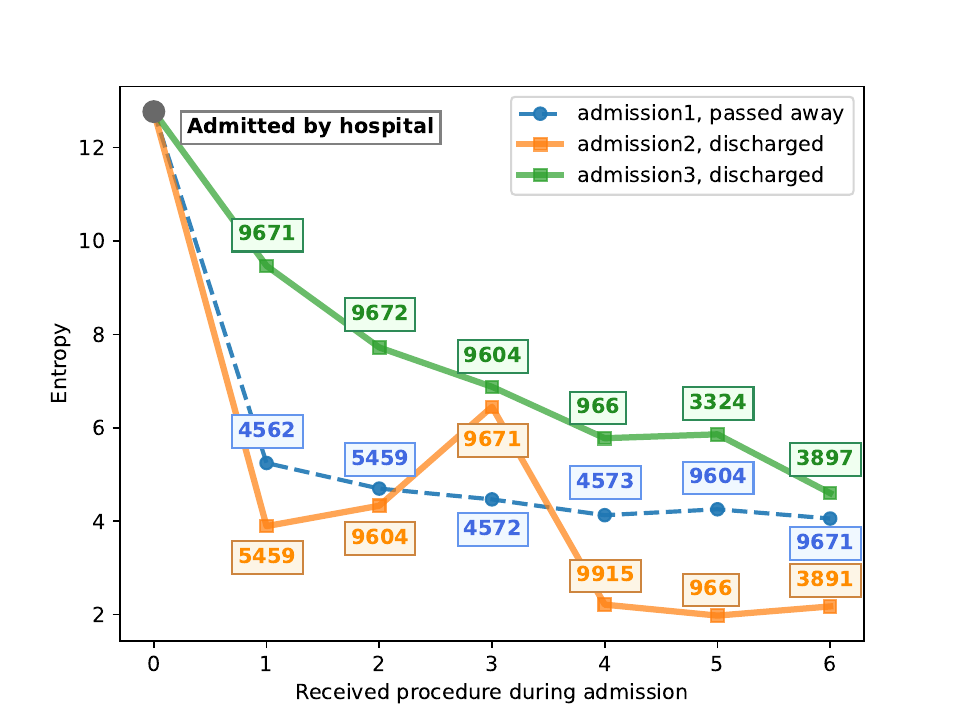}
  \caption{Entropy trends of three admissions with sepsis diagnosis.}
  \label{fig:3case}
\end{figure*}

Figure~\ref{fig:3case} presents the entropy trends of three cases with sepsis diagnosis, each undergoing six procedures. Two cases were discharged from the hospital, and one passed away. These cases were selected from the MIMIC-IV dataset due to their similar final diagnoses and received procedures. The entropy trends are depicted for each admission: Admission \#1 (blue dotted line), Admission \#2 (orange line), and Admission \#3 (green line).

In Admission \#1, a significant drop in entropy is observed after the first procedure, "4562 - Other partial resection of small intestine." However, subsequent procedures do not further reduce the entropy significantly. In Admission \#2, the entropy initially drops dramatically after the first procedure, "5459 - Other lysis of peritoneal adhesions," but rises after procedures 9604 and 9671, before decreasing again following procedure 9915. In contrast, Admission \#3 shows a gradual decrease in entropy, indicating a more consistent reduction in uncertainty with each procedure.

\section{Discussion}

In this study, the focus was on assessing a framework designed to measure and manage medical entropy during various stages of hospital admission. The findings suggest that the framework has potential in quantifying and managing uncertainty in clinical decision-making, adapting to different stages of a patient's hospital stay.

The analysis of trends in Figure~\ref{fig:point_l5} demonstrates the effectiveness of the proposed entropy quantification method in modeling changes in patient conditions. Among the three entropy trends, the trend with all cases is the smoothest. Comparing the differences between the two diagnosis ICD-9 codes, "78650", representing “chest pain” doesn't decline as much as "41401" which denotes "Coronary atherosclerosis of native coronary artery" does. Moreover, "78650" has a more significant standard deviation. This could be attributed to the fact that the conditions related to chest pain are sometimes more uncertain than coronary artery diseases. In addition, the boxplot showing the entropies of the first procedure of "78650" seems to be very concentrated since more than 99\% of the physicians' order an electrocardiogram, which is known as the fastest and most straightforward test to evaluate heart conditions.

\begin{table}[]
\centering
\caption{Entropy of most frequent first N=2 procedures}
\label{tab:2-gram}
\scalebox{0.6}{
\begin{tabular}{@{}cccccc@{}}
\toprule
\multirow{2}{*}{\#} &
  \multirow{2}{*}{\textbf{ICD-9 code}} &
  \multirow{2}{*}{\textbf{Cases}} &
  \multirow{2}{*}{\textbf{Frequency}} &
  \multicolumn{2}{c}{\textbf{Entropy After Receiving}} \\
   &           &      &        & 1st procedure & 2nd procedure \\ \midrule
1  & 0066 3607 & 2920 & 18.72‰ & 9.8504        & 7.3538        \\
2  & 8952 8938 & 1507 & 9.66‰  & 9.9137        & 7.3577        \\
3  & 8952 8744 & 1323 & 8.48‰  & 9.9196        & 7.4492        \\
4  & 3722 8856 & 1316 & 8.44‰  & 10.3795       & 4.5525        \\
5  & 0066 3606 & 930  & 5.96‰  & 9.8362        & 8.8169        \\
6  & 9671 9604 & 925  & 5.93‰  & 10.2844       & 9.3869        \\
7  & 8938 8952 & 867  & 5.56‰  & 7.0969        & 7.1663        \\
8  & 3950 3990 & 696  & 4.46‰  & 9.8251        & 9.3711        \\
9  & 7569 734  & 631  & 4.05‰  & 6.3829        & 5.8259        \\
10 & 9604 9671 & 598  & 3.83‰  & 10.4688       & 9.5779        \\ \bottomrule
\end{tabular}
}
\end{table}

\begin{table}[]
\centering
\caption{Entropy of most frequent first N=3 procedures}
\label{tab:3-gram}
\scalebox{0.6}{
\begin{tabular}{@{}ccccccc@{}}
\toprule
\multirow{2}{*}{\#} &
  \multirow{2}{*}{\textbf{ICD-9 code}} &
  \multirow{2}{*}{\textbf{Cases}} &
  \multirow{2}{*}{\textbf{Frequency}} &
  \multicolumn{3}{c}{\textbf{Entropy After Receiving}} \\
   &                &     &      & 1st procedure & 2nd procedure & 3rd procedure \\ \midrule
1  & 8952 8744 8938 & 989 & 6.3‰ & 9.9174        & 7.4492        & 6.9336        \\
2  & 0066 3607 3722 & 887 & 5.7‰ & 9.7971        & 7.3538        & 4.3214        \\
3  & 8952 8938 8744 & 769 & 4.9‰ & 9.9189        & 7.3577        & 7.1235        \\
4  & 0066 3607 0045 & 705 & 4.5‰ & 9.8413        & 7.3538        & 6.0209        \\
5  & 8938 8952 8744 & 489 & 3.1‰ & 7.1029        & 7.1663        & 6.6327        \\
6  & 3612 3615 3961 & 372 & 2.4‰ & 9.0666        & 3.4497        & 1.4552        \\
7  & 3613 3615 3961 & 359 & 2.3‰ & 9.9924        & 4.3351        & 2.0094        \\
8  & 0066 3607 0040 & 332 & 2.1‰ & 9.8252        & 7.3538        & 6.2223        \\
9  & 0066 3606 3722 & 306 & 2.0‰ & 9.8325        & 8.8169        & 5.6517        \\
10 & 0066 3607 0046 & 286 & 1.8‰ & 9.8265        & 7.3538        & 5.6552        \\ \bottomrule
\end{tabular}
}
\end{table}

The findings from Tables~\ref{tab:1-gram}, ~\ref{tab:2-gram}, and \ref{tab:3-gram} provide insights into how medical procedures influence diagnostic uncertainty. For instance, Table\ref{tab:1-gram} highlights differences in uncertainty between procedures like "7569" (related to delivery) and "9671" (used for respiratory failure). The former typically leads to more straightforward diagnoses, while the latter involves more uncertainty due to various unknown factors. Table~\ref{tab:2-gram} examines entropy drops for the first two procedures after admission. Interestingly, procedures "0066" (PTCA) paired with either "3607" (drug-eluting stent) or "3606" (non-drug-eluting stent) show that the framework can differentiate between these combinations. Additionally, the "3722-8856" pair ("Left heart catheterization" and "Coronary arteriography") shows the largest entropy drop, reflecting the detailed information provided by coronary arteriography. Table~\ref{tab:3-gram} extends this analysis to the first three procedures. The entropy for triplet "3612-3615-3961" (critical coronary bypass surgeries) drops significantly, while triplet "8952-8744-8938" (preliminary heart-related exams) shows only a modest decline. This contrast underscores the importance of life-saving procedures in reducing uncertainty. The entropy trends in Figure~\ref{fig:3case} demonstrate varied impacts of procedures on patient outcomes. In Admission \#1, initial entropy reduction reflects the effectiveness of the first procedure, but later procedures contribute little, suggesting the severity of the condition. Admission \#2 shows fluctuating entropy, indicating varying effectiveness of the procedures, with a final drop suggesting more informative interventions. Admission \#3 displays a steady decrease in entropy, indicating consistent reduction in uncertainty with each procedure.

These findings highlight the importance of procedure sequencing and selection in managing complex cases like sepsis, where uncertainty is high. Understanding entropy trends can help physicians make more informed decisions, offering a more transparent and explainable approach to medical practice. The proposed framework, tested in both simulated and real-world scenarios, demonstrates its potential usefulness across diverse medical cases. The use of ICD codes addresses issues like missing values and outliers, promoting standardization and reliability. These insights have implications for enhancing Clinical Decision Support Systems (CDSS) by integrating data-driven AI/ML models. By quantifying medical entropy, the framework may assist clinicians in making robust, timely decisions, particularly in high-uncertainty situations. However, this study has some limitations. The seq2seq model, while effective here, may not be optimal in all clinical scenarios, and further exploration of alternative models is necessary. Additionally, relying on ICD codes may limit the depth of analysis, as they are generated after patient stays. Their use here is primarily for validation.

Future research could leverage Electronic Health Records (EHR) audit logs to provide a more granular measure of entropy throughout hospital visits, capturing real-time decision-making processes. EHR audit logs offer rich, sequential data that could enhance the predictive power and applicability of our framework in clinical environments \citep{adler2020ehr, kim2023classifying, bhaskhar2023clinical}. 

In summary, this study introduces a novel framework for addressing medical uncertainty in clinical settings. While the approach shows promise in various scenarios, further research is needed to explore its broader impact on healthcare, particularly in advancing Clinical Decision Support Systems.

\section{Conclusion}

In conclusion, this study presents an innovative framework for quantifying and managing medical entropy during hospital admissions. The framework effectively quantifies and adapts to the dynamic nature of clinical decision-making, providing a nuanced understanding of patient-specific variables through the use of entropy quantification. Utilizing the MIMIC-IV dataset and focusing on ICD codes, the study demonstrates how the proposed framework can aid clinicians in reducing diagnostic uncertainty, particularly in complex and time-sensitive medical scenarios. While the seq2seq model used has shown promise, the study acknowledges its potential limitations in certain clinical settings and the need for further exploration of alternative predictive models. Overall, this research contributes significantly to the enhancement of Clinical Decision Support Systems, offering a novel approach to handling medical uncertainty and improving patient care outcomes.

\bibliographystyle{unsrt} 
\bibliography{neurips_2024}

\end{document}